# Recognition of Changes in SAR Images Based on Gauss-Log Ratio and MRFFCM

Jismy Alphonse
M.Tech Scholar
Computer Science and Engineering Department
College of Engineering Munnar, Kerala, India

Biju V. G.
Associate Professor
Electronics and Communication Engineering Department
College of Engineering Munnar, Kerala, India

*Abstract*— A modified version of MRFFCM (Markov Random Field Fuzzy C means) based SAR (Synthetic aperture Radar) image change detection method is proposed in this paper. It involves three steps: Difference Image (DI) generation by using Gauss-log ratio operator, speckle noise reduction by SRAD (Speckle Reducing Anisotropic Diffusion), and the detection of changed regions by using MRFFCM. The proposed method is compared with existing methods such as FCM and MRFFCM using simulated and real SAR images. The measures used for evaluation includes Overall Error (OE), Percentage Correct Classification (PCC), Kappa Coefficient (KC), Root Mean Square Error (RMSE), and Peak Signal to Noise Ratio (PSNR). The results show that the proposed method is better compared to FCM and MRFFCM based change detection method.

## I. INTRODUCTION

Image change detection is the process of recognizing regions of change in images of the same scene taken at different times. This is of widespread interest due to a large number of applications such as medical diagnosis, remote sensing, and video surveillance [1]. The images generated by SARs are used here because they are independent of atmospheric and sunlight conditions. The classical change detection method in SAR images can be divided into two steps: DI generation, and DI analysis [2].

The two well-known techniques for DI generation are differencing (subtraction operator) and rationing (ratio operator). In differencing, changes are measured by subtracting the intensity values between the two images and in rationing, changes are obtained by applying a ratio operator on multitemporal images. If the data are corrupted by additive noise, then image differencing is an appropriate measure in change detection. While SAR images are affected by speckle noise (multiplicative noise), image rationing is better adapted than image differencing. Generally log-ratio operator [1] is used as ratio operator in SAR images.

In DI analysis, the changes are detected by separating the image into two classes. The existence of speckle noise in SAR images, make it difficult to separate two classes changed and unchanged. Smoothing filters are usually used to remove noise. Hence smoothing filter is used after DI generation. But by using smoothing filters important image details can be lost, especially boundaries or edges. The smoothing filters remove edge information, while SRAD [3] filter preserves edges as well as enhances edges by inhibiting diffusion across edges and allowing diffusion on either side of the edge. So SRAD is used in ultrasonic and radar imaging applications.

The changed regions are detected in the DI analysis step. The image segmentation process is used in DI analysis. Image segmentation is the partitioning of an image into different regions. There are two conventional methods for image segmentation, the threshold method and the clustering method [4]. The threshold method is supervised, because some essential models are required to search for a best threshold to divide DI into two classes. And the clustering method is unsupervised, therefore it do not need a model. So clustering method is more convenient and feasible. There are two main clustering strategies: hard clustering scheme and fuzzy clustering scheme. The hard clustering methods classify each pixel of the image to only one cluster, but in soft clustering or fuzzy clustering each pixel of the image may belongs to more than one cluster. In fuzzy clustering, fuzzy set theory introduced the idea of partial membership function. One of the most popular fuzzy clustering methods for image segmentation is the fuzzy c-means (FCM) algorithm [5].

The conventional FCM algorithm works well on noise-free images. It is very sensitive to noise, since it does not consider spatial information. To compensate this drawback, FCM algorithm is modified as FCM_S, FCM_S1, FCM_S2, Enhanced Fuzzy C-means Clustering (EnFCM), Fast Generalized Fuzzy C-means Clustering (FGFCM) and Fuzzy Local Information C-means Clustering (FLICM) [6].

The FCM algorithm is also improved as Reformulated Fuzzy Local Information C-means Clustering (RFLICM) [1] and Genetic algorithm based Fuzzy C Mean (GAFCM) algorithm [7]. Finally MRFFCM [4] is generated for clustering, which considers the information about spatial context effectively. In MRFFCM algorithm the MRF energy function is used with the FCM algorithm to improve the efficiency of FCM algorithm. The MRF serves as a suitable tool to introduce information about the mutual influences among image pixels in a powerful way. The MRFFCM algorithm does not improve FCM by modifying the objective function. Instead, it modifies the membership function of FCM algorithm to reduce the effect of speckle noise.

The rest of this paper is organized as follows: Section II describes the existing change detection method based on MRFFCM and Section III describes the proposed method. The datasets and parameters used are presented in Section IV. The results and discussion on synthetic and real SAR images are described in Section V. Finally, conclusions are drawn in Section VI.

## II. CHANGE DETECTION METHOD BASED ON MRFFCM

The Fig. 1 shows the block diagram of change detection in SAR images using MRFFCM algorithm [4]. Let $I_1$ and $I_2$ are considered as two images: $I_1 = \{I_1(i,j), 1 \leq i \leq m, 1 \leq j \leq n\}$, and $I_2 = \{I_2(i,j), 1 \leq i \leq m, 1 \leq j \leq n\}$, taken over the same geographical area at two different times $t_1$ and $t_2$. The change detection involves two steps: generation of DI from multitemporal images and analysis of DI. The DI is generated using log-ratio operator and the analysis of DI was done using MRFFCM.

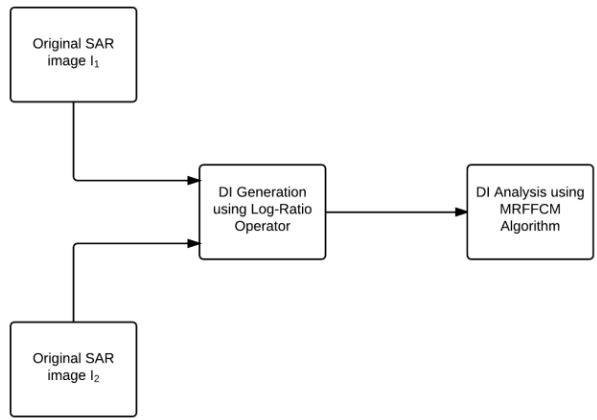

Figure 1. Block diagram for the MRFFCM based change detection approach

The log-ratio operator $X_l$ can be defined as the logarithm of intensity values of second image to the first image.

$$X_l = \left| \log \frac{X_2}{X_1} \right| \quad (1)$$

The main procedure of MRFFCM is as follows [4].
1) To derive the mean $\mu_i^1$ and the standard deviation $\sigma_i^1$ using K and I method [8]. Initial membership matrix $\{u_{ij}^1\}$ is generated by utilizing original FCM algorithm unmodified [6]. Generate the number matrix $\{n_{i\in\partial j}^1\}$.
2) To establish the energy matrix $\{E_{ij}^k\}$, using the equation,

$$E_{ij} = -\ln(mu_{ij}) + \beta_j(mu_{ij}, n_{ij}) \cdot t_{qj} \cdot n_{ij} \quad (2)$$

where $mu_{ij}$ be the mean partition matrix, $n_{ij}$ be the number matrix, $t_{qj}$ adjusts the sign of term it only concerned with the class that the central pixel belongs to and the location of it.
It is defined as:

$$t_{qj} = \text{sgn}(u_{qj} - 0.5) \quad (3)$$

3) To compute the pointwise prior probabilities of the MRF using the energy function as:

$$\pi_{ij}^k = \frac{\exp(-E_{ij}^k)}{\exp(-E_{uj}^k) + \exp(-E_{cj}^k)} \quad (4)$$

4) To compute the conditional probability $P_i^k$ using mean and standard deviation as:

$$P_i^k(y_j|\mu_i^k, \sigma_i^k) = \frac{1}{\sigma_i^k\sqrt{2\pi}} \exp\left[-\frac{(y_j - \mu_i^k)^2}{2(\sigma_i^k)^2}\right] \quad (5)$$

And to generate the distance matrix $d_{ij}^k$, which is defined as the negative logarithm of conditional probability $\{P_i^k\}$

$$d_{ij}^k = -\ln[P_i^k(y_j|\mu_i^k, \sigma_i^k)] \quad (6)$$

5) To compute the objective function, this is the sum of sum of square of partition matrix and square of distance matrix:

$$J_{ij}^k = \sum_{i=u,c} \sum_{j\in I_x} (u_{ij}^k)^2 (d_{ij}^k)^2 \quad (7)$$

If the distance between objective function and previous objective function is less than or equal to threshold, then the algorithm stops and return partition matrix. That is

$$|J_{ij}^k - J_{ij}^{k-1}| \leq \delta \quad (8)$$

In case of convergence, exit and output $\{u_{ij}^k\}$; otherwise go to 6.

6) To compute the new membership matrix using distance matrix and pointwise prior probability matrix

$$u_{ij}^{k+1} = \frac{\pi_{ij}^k \exp(-d_{ij}^k)}{\pi_{uj}^k \exp(-d_{uj}^k) + \pi_{cj}^k \exp(-d_{cj}^k)}$$
(9)

7) Update the mean and the standard deviation as $\mu_i^{k+1}$ and $\sigma_i^{k+1}$

$$\mu_i^{k+1} = \frac{\sum_{j \in I_x}(u_{ij}^k y_j)}{\sum_{j \in I_x}(u_{ij}^k)}$$
(10)

$$\sigma_i^{k+1} = \sqrt{\frac{\sum_{j \in I_x}[u_{ij}^k(y_j - \mu_i^{k+1})^2]}{\sum_{j \in I_x} u_{ij}^k}}$$
(11)

The results obtained on MRFFCM based change detection method applied on SAR images shows that the method still requires some improvement for the analysis of noisy SAR images. So a modification of the method is suggested.

### III. PROPOSED METHODOLOGY

The proposed change detection method is a three step process: the generation of the DI, to reduce speckle noise, and the analysis of the DI. It contains three parts: 1) generating the DI using Gauss-log ratio operator; 2) image denoising via SRAD; 3) the DI analysis by MRFFCM algorithm as shown in Fig. 2. These three parts will be detailed as follows:

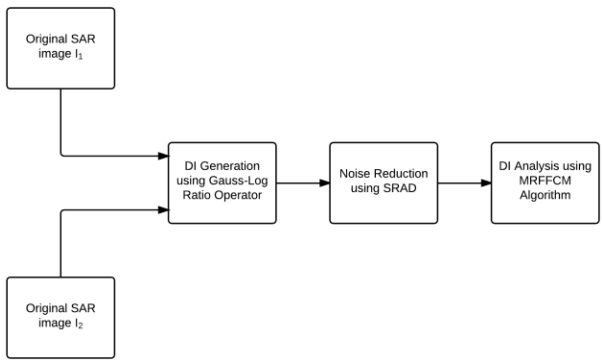

Figure 2. Block diagram for the proposed change detection approach

#### A. Generating the DI using Gauss-log ratio operator

To generate the DI the log-ratio operator is widely used. The ratio image is expressed in a logarithmic scale to enhance low-intensity pixels. But the information of changed portions obtained by the log-ratio image cannot reflect the real changed region completely because of the weakening in the areas of high-intensity pixels [1]. Not only to enhance the real change trend but also suppress the unchanged portions in the DI, in the proposed method to replace log-ratio operator with Gauss-log ratio operator [9]. It considers the relationship between the intensities of local patches of two SAR images.

The Gauss-log ratio operator can be defined as:

$$X_r(i,j) = \sum_{m=-1}^{1} \sum_{n=-1}^{1} |X_{r1}(i+m, j+n) - X_{r2}(i+m, j+n)|$$
(12)

$$X_{r1}(i,j) = log(X_1'(i,j)) * G$$
(13)

$$X_{r2}(i,j) = log(X_2'(i,j)) * G$$
(14)

where $X_1'(i,j)$ and $X_2'(i,j)$ are two patches centered at point $(i,j)$ in SAR images $X_1$ and $X_2$, respectively. G is a rotationally symmetric Gaussian low-pass filter of size $3 \times 3$ with standard deviation 0.5.

#### B. Image denoising via SRAD

The speckle reduction is an important problem in SAR images. It is edge sensitive and multiplicative process. The SRAD [4] is used in the proposed method for speckle noise reduction, which removes the speckle noise without blurring the image. Conventional anisotropic diffusion is the edge-sensitive diffusion for images corrupted with additive noise. Instead, SRAD is the edge-sensitive diffusion for speckled image. So here SRAD is used.

The main procedure of SRAD is as follows.

1) The speckle scale function $q_0(t)$ can be estimated using

$$q_0(t) = \frac{\sqrt{var[z(t)]}}{\overline{z(t)}}$$
(15)

where $var[z(t)]$ and $\overline{z(t)}$ are the intensity variance and mean over a homogeneous area at $t$, respectively.

2) The instantaneous coefficient of variation determined by

$$q(x,y;t) = \sqrt{\frac{\left(\frac{1}{2}\right)\left(\frac{|\nabla I|}{I}\right)^2 - \left(\frac{1}{4^2}\right)\left(\frac{\nabla^2 I}{I}\right)^2}{\left[1 + \left(\frac{1}{4}\right)\left(\frac{\nabla^2 I}{I}\right)\right]^2}}$$
(16)

3) The diffusion coefficient defined as:

$$c(q) = \frac{1}{1 + \frac{[q^2(x,y;t) - q_0^2(t)]}{[q_0^2(t)(1 + q_0^2(t))]}}$$
(17)

4) The SRAD update function can be calculated as:

$$I_{i,j}^{n+1} = I_{i,j}^n + \frac{\nabla t}{4} d_{i,j}^n$$
(18)

where

$$d_{i,j}^n = c_{i,j}^n\left(I_{i-1,j}^n - I_{i,j}^n\right) + c_{i+1,j}^n\left(I_{i+1,j}^n - I_{i,j}^n\right) \\ + c_{i,j}^n\left(I_{i,j-1}^n - I_{i,j}^n\right) + c_{i,j+1}^n\left(I_{i,j+1}^n - I_{i,j}^n\right)$$

(19)

and $\nabla t = 0.5$.

*C. The DI analysis by MRFFCM*

For the analysis of DI the MRFFCM [4] algorithm is used in the proposed change detection method. It improves the conventional FCM by modifying the membership of each pixel. The spatial context is considered in MRFFCM algorithm. It classifies the changed and unchanged region effectively than the existing methods.

### IV. DATASETS AND PARAMETERS USED FOR EVALUATION

In order to assess the effectiveness of the proposed approach, in this section, synthetic datasets with different levels of noises and three sets of real SAR images are considered. And to introduces some parameters for evaluation criteria.

*A. Datasets*

Five set of synthetic images are generated for evaluation. Four additive salt&pepper noise levels with different values of noise densities (d) such as 0.05, 0.10, 0.15, and 0.20 are added to each set of synthetic images. And four multiplicative speckle noise levels with different values of variances (v) such as 0.10, 0.20, 0.30, and 0.40 are added to each set of synthetic images.

Three set of real SAR images are also used for evaluation [10]. First set contain two SAR images over an area near the city of Bern, Switzerland acquired by ERS−2 (European Remote Sensing) satellite in April and May 1999. Second set hold two SAR images over the city of Ottawa, Canada acquired by RADARSAT satellite in May and August 1997. And third set comprise two SAR images over the region of the Yellow River Estuary, China acquired by RADARSAT−2 satellite in June 2008 and June 2009.

*B. Parameters*

For the synthetic images the following parameters were calculated.

*1) Overall Error (OE)*

The overall error is defined as the percentage of sum of false positive and false negative, which can be calculated as:

$$OE = \left[\frac{(FP + FN)}{N}\right] * 100$$

(20)

where FP (False Positive) is defined as the number of pixels belonging to the unchanged class but falsely classified as changed class, FN (False Negative) denote the number of pixels belonging to the changed class but falsely classified as unchanged class, and N be the number of entire pixels.

$$N = TP + TN + FP + FN \quad (21)$$

TP (True Positive) and TN (True negative) represent the number of changed pixels and unchanged pixels respectively, which are calculated as follows:

$$\begin{cases} TP = N_c - FP \\ TN = N_u - FN \end{cases} \quad (22)$$

$N_c$ is the actual number of pixels belonging to the changed class, and $N_u$ is the actual number of pixels belonging to the unchanged class [4].

*2) Percentage Correct Classification (PCC)*

The percentage correct classification is the percentage of sum of true positive and true negative [4], which is calculated as:

$$PCC = \left[\frac{(TP + TN)}{N}\right] * 100$$

(23)

*3) Kappa Coefficient (KC)*

The Kappa Coefficient [4] is defined as:

$$KC = \frac{PCC - PRE}{1 - PRE}$$

(24)

where

$$PRE = \frac{[(TP + FN).N_c + (FP + TN).N_u]}{N^2}$$

(25)

The value of *KC* ranges from 0 to 1.

*4) Root Mean Square Error (RMSE)*

The Root Mean Square Error (RMSE) can be calculated as:

$$RMSE = \sqrt{MSE} \quad (26)$$

where

$$MSE = \frac{1}{mn}\sum_{i=0}^{m-1}\sum_{j=0}^{n-1}[I(i,j) - K(i,j)]^2$$

(27)

*K* is the segmented output image and *I* be the noise-free $m \times n$ reference image. The value of *RMSE* can be varying from 0 to infinity [11].

5) *Peak Signal to Noise Ratio (PSNR)*

The parameter Peak Signal to Noise Ratio (*PSNR*) can be defined as:

$$PSNR = 10\, log\left(\frac{MAX_I^2}{MSE}\right) \qquad (28)$$

where $MAX_I$ is the maximum possible pixel value of image and *MSE* is the mean square error. The value of *PSNR* ranges from 0 to 99 [12].

## V. Results And Discussion

The efficiency and robustness of the proposed change detection method is compared with existing MRFFCM based method. The results of MRFFCM based method is obtained based on block diagram shown in Fig. 1 and the results of proposed method is obtained based on block diagram shown in Fig. 2. These methods are applied on synthetic images corrupted by different levels of noises and real SAR images.

The original synthetic image is shown in Fig. 3(a) and the changed image is shown in Fig. 3(b). The difference of Fig. 3(a) and 3(b) is shown in Fig. 3(c), which is generated by differencing Fig. 3(a) and 3(b). Fig. 3(c) is known as difference image or reference image. The salt&pepper noise (d=0.05) is added to Fig. 3(a) and 3(b) which are shown in Fig. 3(d) and 3(e), respectively. Fig. 3(f) and 3(g) depict the change detection results of MRFFCM and proposed method, respectively.

Each level of salt&pepper noise is added five times to synthetic image and tested, the average value is taken as the result. The testing results are shown in Table I, which shows that the performance of proposed method is better than MRFFCM.

According to the findings on the synthetic data in Table I, the PCC, KC and PSNR values between the MRFFCM based change detection method and the proposed method are increased by 7.540%, 0.465 and 23.029dB, respectively. The OE and RMSE values are decreased by 7.539% and 1.209, respectively. Better change detection method obtains highest PCC, KC and lowest OE values. The highest values of PCC, KC and lowest value of OE are achieved in proposed method, which are about 99.831%, 0.985 and 0.115% respectively. So the proposed method is better compared to the MRFFCM based method. The proposed method also attained the lowest RMSE value and highest PSNR value, which are about 0.047 and 74.742dB, respectively.

TABLE I. Values of the Evaluation Criteria of the Synthetic Dataset which are Corrupted by different levels of Salt&Pepper Noise

| Salt&Pepper (d=0.05) | | | | | |
|---|---|---|---|---|---|
| | OE (%) | PCC (%) | KC | RMSE | PSNR (dB) |
| MRFFCM | 1.643 | 98.357 | 0.815 | 0.298 | 58.663 |
| Proposed | 0.115 | 99.885 | 0.985 | 0.047 | 74.742 |
| Salt&Pepper (d=0.10) | | | | | |
| | OE (%) | PCC (%) | KC | RMSE | PSNR (dB) |
| MRFFCM | 5.261 | 94.739 | 0.559 | 0.879 | 49.245 |
| Proposed | 0.169 | 99.831 | 0.978 | 0.063 | 72.179 |
| Salt&Pepper (d=0.15) | | | | | |
| | OE (%) | PCC (%) | KC | RMSE | PSNR (dB) |
| MRFFCM | 9.698 | 90.302 | 0.383 | 1.596 | 44.069 |
| Proposed | 0.240 | 99.759 | 0.969 | 0.083 | 69.759 |
| Salt&Pepper (d=0.20) | | | | | |
| | OE (%) | PCC (%) | KC | RMSE | PSNR (dB) |
| MRFFCM | 14.418 | 85.582 | 0.271 | 2.366 | 40.652 |
| Proposed | 0.338 | 99.666 | 0.957 | 0.109 | 68.067 |

The original synthetic image is shown in Fig. 4(a) and the changed image is shown in Fig. 4(b). The difference of Fig. 4(a) and 4(b) is shown in Fig. 4(c), which is generated by differencing Fig. 4(a) and 4(b). Fig. 4(c) is known as difference image or reference image. The speckle noise (v=0.20) is added to Fig. 4(a) and 4(b) which are shown in Fig. 4(d) and 4(e), respectively. Fig. 3(f) and 3(g) depict the change detection results of MRFFCM and proposed method, respectively.

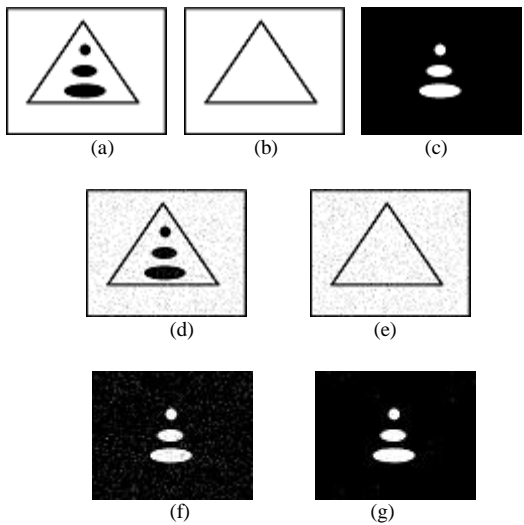

Figure 3. Synthetic images corrupted by salt&pepper noise. (a) Original image I1. (b) Changed image of I1, I2. (c) Reference image. (d) Image I1 with salt&pepper noise(d=0.05). (e) Image I2 with salt&pepper noise(d=0.05). (f) Change detected image using MRFFCM. (g) Change detected image using proposed method.

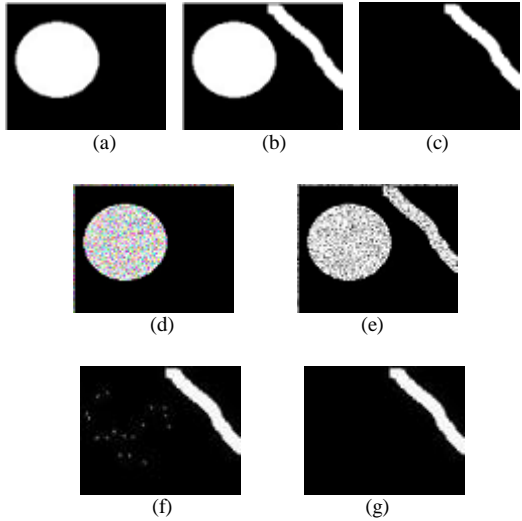

Figure 4. Synthetic images corrupted by speckle noise. (a) Original image I1. (b) Changed image of I1, I2. (c) Reference image. (d) Image I1 with speckle noise(v=0.20). (e) Image I2 with speckle noise(v=0.20). (f) Change detection image of MRFFCM. (g) Change detection image of proposed method.

Each level of speckle noise is added five times to synthetic image and tested, the average value is taken as the result. The testing results are shown in Table II, which shows that the performance of proposed method is better than MRFFCM.

According to the findings on the synthetic data in Table II, the proposed method achieved the highest PCC, KC, PSNR and lowest OE, RMSE values as compared to the MRFFCM based method. The PCC, KC and PSNR values between the MRFFCM based change detection method and the proposed method are increased by 1.726%, 0.109 and 20.318dB, respectively. The OE and RMSE values are decreased by 1.726% and 0.263, respectively. Better change detection method obtains highest PCC, KC and lowest OE values. The highest values of PCC, KC and lowest value of OE are achieved in proposed method, which are about 100%, 1 and 0% respectively. So the proposed method is better compared to the MRFFCM based method. The proposed method also attained the lowest RMSE value and highest PSNR value, which are about 0 and 99dB, respectively.

TABLE II. VALUES OF THE EVALUATION CRITERIA OF THE SYNTHETIC DATASET WHICH ARE CORRUPTED BY DIFFERENT LEVELS OF SPECKLE NOISE

| Speckle (v=0.10) | | | | | |
|---|---|---|---|---|---|
| | OE (%) | PCC (%) | KC | RMSE | PSNR (dB) |
| MRFFCM | 0 | 100 | 1 | 0 | 99 |
| Proposed | 0 | 100 | 1 | 0 | 99 |
| Speckle (v=0.20) | | | | | |
| | OE (%) | PCC (%) | KC | RMSE | PSNR (dB) |
| MRFFCM | 0.214 | 99.786 | 0.983 | 0.058 | 72.897 |
| Proposed | 0 | 100 | 1 | 0 | 99 |
| Speckle (v=0.30) | | | | | |
| | OE (%) | PCC (%) | KC | RMSE | PSNR (dB) |
| MRFFCM | 2.349 | 97.651 | 0.839 | 0.379 | 56.544 |
| Proposed | 0.005 | 99.995 | 0.999 | 0.006 | 92.969 |
| Speckle (v=0.40) | | | | | |
| | OE (%) | PCC (%) | KC | RMSE | PSNR (dB) |
| MRFFCM | 4.716 | 95.284 | 0.712 | 0.700 | 51.238 |
| Proposed | 0.369 | 99.630 | 0.971 | 0.081 | 69.980 |

The change detection results of two SAR images over the city of Bern, Switzerland in April and May 1999 are shown in Fig. 5. Fig. 6 illustrates the change detection results of two SAR images over the city of Ottawa, Canada in May and August 1997. And the change detection results of two SAR images over the region of the Yellow River, China in June 2008 and June 2009 are shown in Fig. 7.

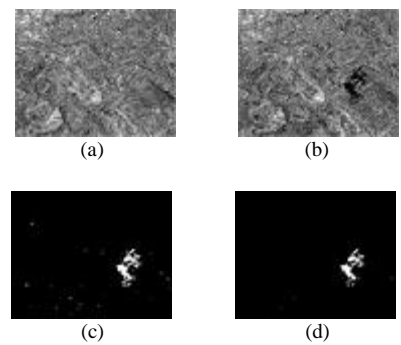

Figure 5. Images of Bern. (a) Original image I1, acquired in April 1999. (b) Original image I2, acquired in May 1999. (c) Change detection image generated by MRFFCM. (d) Change detection image generated by proposed method.

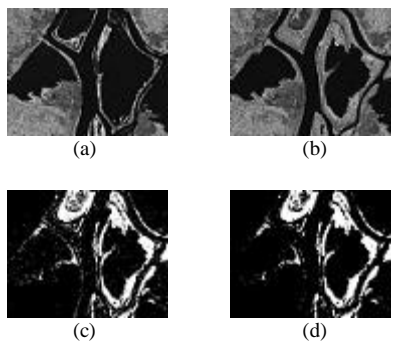

Figure 6. Images of Ottawa. (a) Original image I1, acquired in May 1997. (b) Original image I2, acquired in August 1997. (c) Change detection image generated by MRFFCM. (d) Change detection image generated by proposed method.

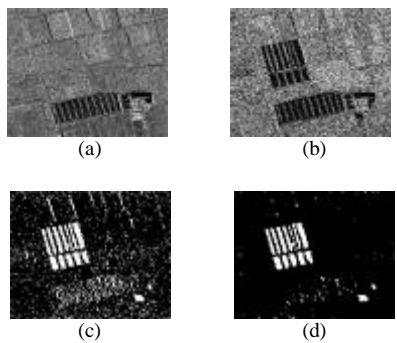

Figure 7. Images of Yellow River. (a) Original image I1, acquired in June 2008. (b) Original image I2, acquired in June 2009. (c) Change detection image generated by MRFFCM. (d) Change detection image generated by proposed method.

## VI. CONCLUSION

In this paper, a new method for change detection in SAR images has been presented. This approach is based on the Gauss-log ratio operator and MRFFCM algorithm. In the DI generation step, two multi temporal images are compared to generate a Gauss log-ratio image that contains explicit information on changed areas. Then SRAD is applied to reduce the effect of speckle noise. Finally, MRFFCM algorithm is applied to detect the changed and unchanged regions. It focuses on modifying the membership function of FCM algorithm instead of modifying the objective function. The proposed method is tested on both synthetic and real images. The proposed change detection method is compared against the existing MRFFCM based method, which uses log-ratio operator for DI generation. The evaluation result shows that the proposed approach is better compared with the MRFFCM based method.